\documentclass[10pt, a4paper]{article}

\usepackage{lrec-coling2024} % this is the new style
\usepackage{amsmath}
\usepackage{amsfonts}
\usepackage{booktabs,siunitx}
\usepackage{subfigure}
\usepackage{todonotes}

\usepackage{siunitx}
\usepackage{booktabs}

\title{Improving the Robustness of Large Language Models via Consistency Alignment}
\name{Yukun Zhao\textsuperscript{\rm 1,2}, Lingyong Yan\textsuperscript{\rm 2}, Weiwei Sun\textsuperscript{\rm 1}, Guoliang Xing\textsuperscript{\rm 2}, Chong Meng\textsuperscript{\rm 2} \\
\large{\textbf{Shuaiqiang Wang}\textsuperscript{\rm 2},  \textbf{Zhicong Cheng}\textsuperscript{\rm 2},  \textbf{Zhaochun Ren}\textsuperscript{\rm 3\textasteriskcentered}\thanks{\textasteriskcentered ~~Co-corresponding authors.}, \textbf{Dawei Yin}\textsuperscript{\rm 2\textasteriskcentered}} }

\address{\textsuperscript{\rm 1}Shandong University, Qingdao, China, \textsuperscript{\rm 2}Baidu Inc., Beijing, China \\
\textsuperscript{\rm 3}Leiden University, Leiden, The Netherlands \\
\texttt{\{zhaoyukun02,yanlingyong\}@baidu.com,sunnweiwei@gmail.com}\\
\texttt{\{xingguoliang,mengchong01,wangshuaiqiang,chengzhicong01\}@baidu.com} \\
\texttt{z.ren@liacs.leidenuniv.nl},  \texttt{yindawei@acm.org}
}

\abstract{
Large language models (LLMs) have shown tremendous success in following user instructions and generating helpful responses. Nevertheless, their robustness is still far from optimal, as they may generate significantly inconsistent responses due to minor changes in the verbalized instructions. Recent literature has explored this inconsistency issue, highlighting the importance of continued improvement in the robustness of response generation. However, systematic analysis and solutions are still lacking. In this paper, we quantitatively define the inconsistency problem and propose a two-stage training framework consisting of instruction-augmented supervised fine-tuning and consistency alignment training. The first stage helps a model generalize on following instructions via similar instruction augmentations. In the second stage, we improve the diversity and help the model understand which responses are more aligned with human expectations by differentiating subtle differences in similar responses. The training process is accomplished by self-rewards inferred from the trained model at the first stage without referring to external human preference resources.
We conduct extensive experiments on recent publicly available LLMs on instruction-following tasks and demonstrate the effectiveness of our training framework.
 \\ \newline  }

\begin{document}

\maketitleabstract

\section{Introduction}

Large language models(LLMs) are now regarded as one of the most advancing fields in artificial intelligence researches~\cite{OpenAI2023GPT4TR, vicuna2023, taori2023alpaca,touvron2023llama}.
By sufficiently pre-training on massive textual corpus, and carefully fine-tuning and aligning on high-quality instruction-following data,
LLMs have demonstrated remarkable capabilities, e.g. understanding human instructions and generating helpful responses~\cite{wei2021finetuned, ouyang2022training, dong2023raft, rafailov2023direct, yuan2023rrhf}.
%And the vast potential of LLMs has attracted many downstream applications in different areas~\cite{xxx}.
% Large language models (LLMs)~\cite{OpenAI2023GPT4TR, vicuna2023, taori2023alpaca,touvron2023llama} have achieved remarkable success for various natural language processing (NLP) tasks\cite{xxx}. They have been shown great potential to understand human instructions and generate helpful and desired responses via instruction tuning~\cite{wei2021finetuned, ouyang2022training, dong2023raft, rafailov2023direct, yuan2023rrhf}.

\begin{figure}[t]
\vspace{0.0cm}
    \begin{centering}
    \includegraphics[width=1.0\linewidth]{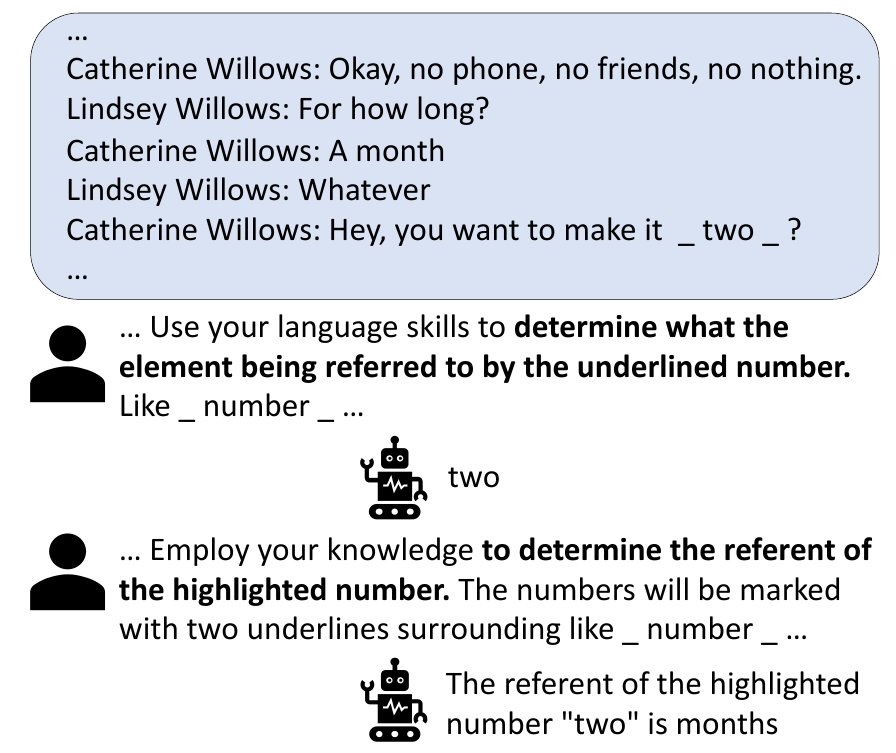}
    \end{centering}
    \vspace{-0.4cm}
    \caption{GPT-4 generates inconsistent responses for the identical task.}
    \vspace{-0.1cm}
    \label{fig:intro_sample}
    \vspace{-0.4cm}
\end{figure}

However, the robustness of current LLMs, even those leading ones, is still far from promising in recent literature~\cite{gu2022robustness,sun2023evaluating,liang2023exploring}.
A commonly observed phenomenon is the inconsistency problem when they respond to distinct but semantically equivalent instructions.
We list an example shown in Figure~\ref{fig:intro_sample}: we see GPT-4 returns inconsistent answers to the same task "the referent of the number".
Such inconsistency problem reflects the inherent flaws of LLMs to some extent and hinders their practical applications. 

Recent work explores the inconsistency problem~\cite{gu2022robustness,sun2023evaluating,li2023instruction,liang2023exploring}.
For instance, they find the LLMs may generate inconsistent responses due to the different verbalized instructions~\cite{li2023instruction}, data distribution shift~\cite{li2023robust}, or even discrepancies in instruction formats~\cite{gu2022robustness}.
Based on these observations, \citet{li2023robust} and \citet{liang2023exploring} propose to optimize the instruction to identify the optimal task instruction that elicits the best performance for LLMs. Nevertheless, there is an absence of quantitative analysis of the current state, along with a systemic solution to improve the instruction-tuned LLMs.
% \todo[]{hard to understand} 
% Nevertheless, these studies mainly focus on how to detect and avoid the inconsistency problem in different LLMs, rather than systemically alleviating it.

In this paper, we first quantitatively analyze the generation robustness of current LLMs in terms of our consistency metrics. We then propose a novel training framework for LLMs via consistency alignment to mitigate the inconsistency problem in current LLMs.
Concretely, our training framework sequentially performs the following two training stages: instruction augmented supervised fine-tuning and response consistency alignment.
(1) In the augmented supervised fine-tuning (SFT) stage, we first paraphrase the original instruction in the SFT dataset and then pair each paraphrased instruction with the original response to form a new augmented training sample.
Finally, all augmented training samples are then added to the SFT dataset to fine-tune the LLMs.
(2) In the consistency alignment stage, we feed the paraphrased instructions to LLMs to generate candidate responses, and then construct <good, bad> response pairs where each response is individually evaluated by the consistency score.
Finally, we optimize the LLMs to directly learn the preferences through an offline training algorithm~\cite{yuan2023rrhf}.
%Moreover, to effectively assess the consistency degree of each LLM on any testing dataset, we propose a new consistency score.

We conduct extensive experiments on publicly available models including Vicuna-7B, Vicuna-13B, Llama2-7B, and Llama2-13B on the instruction-following tasks.
The experimental results show that by explicitly adding consistency self-alignment, these LLMs can obtain robustness improvements and generalize better on following instructions.

%To the best of our knowledge, we are the first to systemically analyze the robustness of the large language models, and propose an integrated training framework to address this problem.
Our contributions are as follows:
\begin{enumerate}
    \item We propose an integrated training framework to enhance the robustness of LLMs. 
    %Our training framework can be used as an add-on for further improvement of an existing trained model.
    \item We propose to utilize self-rewards to improve the performance of a large language model without referring to external human preference resources or external reward models. 
    \item We conduct extensive experiments to verify the effectiveness of our training framework method across several public LLMs.
    % \item Our ablation study reveals that the performance can be improved through an appropriate amount of augmented instructions and more accurate and fine-grained self-rewards.
\end{enumerate}

\section{Related Work}
\paragraph{Instruction Tuning.}
% \todo[] {abbreviate}
In order to help LLMs understand the instructions and generate human expected responses, recent work~\cite{ouyang2022training, vicuna2023,taori2023alpaca, wei2021finetuned} employ instruction fine-tuning on the pre-trained models to help them follow user instructions. \citet{ouyang2022training} propose to optimize the fine-tuned model (policy model) with PPO to learn human preference. \citet{dong2023raft} propose a reward-ranked fine-tuning method, which selects the top n model outputs using an existing reward model to fine-tune foundational LLMs. \citet{rafailov2023direct} proposes to directly optimize preferences between two responses given a specific instruction, which implicitly optimizes the same objective as existing PPO algorithms. \citet{song2023preference} and \citet{ziegler2019fine} propose similar methods to further fine-tune the LLMs utilizing the ranked response pairs that align with human preference. \citet{zhao2022calibrating} and \citet{zhao2023slic} calibrate the sequence likelihood by sampling generated candidates and making the candidates align with the references in latent space using various ranking loss. \citet{yuan2023rrhf} continue to optimize a bigger model LLAMA-7B and propose RRHF which is a similar method as described above.

%Due to the substantial resources demands during the fine-tuning process, many parameter efficient training methods, such as prefix tuning~\cite{li2021prefix}, prompt tuning~\cite{lester2021power}, LoRA~\cite{hu2021lora}, AdaLoRA~\cite{zhang2023adaptive} and QLoRA~\cite{dettmers2023qlora} have been proposed.

%性能训练

%觉醒到相关的

%alpaca, vicuna, llama, flan, 
%FINETUNED LANGUAGE MODELS ARE ZERO-SHOT LEARNERS

\paragraph{Prompting.} Prompting is attractive for its simplicity to improve alignment for the LLMs by using few samples or suitable instructions~\cite{brown2020language,jiang2021can}. \citet{wei2022chain} propose Chain-of-thought (CoT) to improve reasoning abilities. Their successors \cite{zhou2022least} propose least-to-most prompting to solve complex reasoning tasks. \citet{wang2022self} and \citet{si2022prompting} propose to utilize the self-consistency between the sampled answers and choose the most frequent one as the final answer.

\paragraph{Instruction Data.}
An intuitive start point is to collect a substantial array of diverse and heterogeneous NLP tasks from existing benchmarks~\cite{wei2021finetuned,longpre2023flan,wang2022super} for instruction-tuning. Then \citet{conover2023free}, \citet{kopf2023openassistant} and \citet{vicuna2023} collect crowd-sourcing human-written instructions. \citet{wang2022self}, \citet{yu2023large} and \citet{xu2023wizardlm} prompt LLMs to generate large-scale, diverse and more complex instructions automatically. Recent work~\cite{zhou2023lima, cao2023instruction,  chen2023alpagasus, jiang2023lion} focus on generating or selecting high-quality and representative instructions to improve the instruction tuning performance. 

%collect diverse nlp tasks, FLAN, natural instructions
%collect crownd-sourcing task
% prompt llms to generate high qualiy, diverse
% identify high quality instructions

\paragraph{Robustness on Instruction-following.}
Recent work~\cite{gu2022robustness, liang2023exploring} have explored that the manipulated instructions would degrade the performance of instruction-tuned LLMs. \citet{li2023instruction} evaluate the instruction-following abilities of LLMs through different verbalizations and emphasize the need for continued improvement on instruction-following abilities. \citet{li2023robust} consider the distribution shift between the seen training data and the unseen test data and propose an ensemble method to derive optimal instructions to elicit the performance on the unseen data group. \citet{sun2023evaluating} reveal that instruction-tuned LLMs are sensitive to instruction re-phrasings, and propose soft prompts by transferring the manipulated instruction to the optimal ones to alleviate this issue. All of the previous work lacks a quantitative analysis of the current state in the robustness of the generations, along with corresponding solutions.
% 1. 定义robustness， 在相似的instruction上输出一致。
% 在paraphrased的结果中，report 和原始回答不一致的比例, 如 3/10
% 相似的instructions 是指xxx
% 输出一致是指的 xx
% 2. 不一致分析，1）没理解任务； 2) 回答错误

\section{Robustness on Instruction Following}
\label{sec:definition}

% \subsection{Evaluation Metric of Robustness}
% \todo[]{Huh???}The response consistency to varying textual instructions that describe the same semantics or user intentions is a vital aspect of the robustness in the context of Large Language Models (LLMs).
Given the multifarious ways in which natural language conveys identical semantics, it is crucial for LLMs to maintain answer consistency across various verbalized questions or instructions. We characterize the robustness in terms of the answer consistency and analyze the current state of LLMs in this regard.

\paragraph{Definition of Consistency}

We denote \(Q\) as a potential space, representing all conceivable linguistic paraphrases conveying equivalent semantic content or user intent. Let \(Y: Q \rightarrow \mathcal{P}(Y)\) be a function mapping each question in \(Q\) to a probability distribution over the possible responses in \(Y\).
Given this, the consistency of an LLM $\mathcal{R}$ is then defined as the expected consistency between the model’s responses to any two elements from \(Q\):
\begin{equation}
    \mathcal{R} = \mathbb{E}_{q_i, q_j \in Q} \left[ \mathbb{E}_{y_i \sim Y(q_i), y_j \sim Y(q_j)}[\text{{sim}}(y_i, y_j)] \right]
\end{equation}

where $\text{sim} (y_i, y_j) \in [0, 1]$ is a function measuring the consistency between two responses in \(Y\). A higher value of $\mathcal{R}$ denotes greater robustness. It demonstrate the model's ability to maintain consistency across diverse and potentially infinite linguistic representations in \(Q\), despite the inherent response variability.

%\paragraph{Definition of Similarity Function}

For the similarity function, \(\text{{sim}}\),
the feasible approaches include measuring the similarity of the response embeddings~\cite{zhang2019bertscore,pillutla2021mauve}, or exploiting LLMs to check whether the two answers are the same or contradicted~\cite{kadavath2022language}. Due to the limitations of semantic similarity based on embeddings and the ability of LLMs that align with human intent better in such tasks, we choose to use LLMs to assess the consistency of responses. That is, we prompt the LLMs to determine whether two responses are similar or contradicted and the instruction is shown in Table~\ref{tab:NLI_prompt}. Then $sim(y_i, y_j) \in \{0,1\}$ is obtained by inferring the generated contents.

In particular, we formally define Consistency Rate $CR$ and Maximum Consistency Rate $MCR$ as the consistency metrics. The first one is an indicator of the consistency rate between any two answers under $Q$, 
\begin{equation}
    CR = \frac{1}{|Q|} \sum_{Q_k \in Q} \sum_{y_i\in Y_k} \sum_{y_j\in Y_k, j\neq i} \frac{sim(y_i, y_j)}{ \tbinom{|Y_k|}{2}}
\end{equation}
where $\tbinom{|Y_k|}{2}$ is the number of 2-combinations for the responses $Y_k$ under the same input. We use Maximum Consistency Rate $MCR$ to report the rate of the maximum consistent answers among the all generated answers on the $Q$ Tasks. 
\begin{equation}
    MCR = \frac{1}{|Q|} \sum_{Q_k \in Q} \frac{|\Omega_{k}^{max}|}{|Y_k|} 
\end{equation}
where $|\Omega_{k}^{max}| = \arg_{j} |\max \Omega_{j}|$ and $\Omega_{j}$ is a cluster of consistent answers under the same input.

\begin{table}[h]
%\centering
\small
% \vspace{-0.1cm}
\begin{tabular}[c]{p{7.25cm}}
\toprule
\texttt{Determine whether answer "A" is the same or contradicted with the answer "A Reference" for the question "Q". For the tasks with fixed answers, if the two answers are exactly the same you give "same", otherwise, you give "Contradicted" as the output. For free-form generation tasks, you need to check whether the answer "A" is an expected generated title, question, data-to-text description, or summarization, etc., as the answer "A Reference". If the two answers describe a similar meaning you give "Same", otherwise, you give "Contradicted" as the output.}  \\
\bottomrule
\end{tabular}
\caption{The instruction for determining whether two responses are consistent.}
\label{tab:NLI_prompt}
\vspace{-0.2cm}
\end{table}

% In particular, we use the $CR$ and $MCR$ metrics to assess the robustness of the current LLMs.

% One feasible approach to quantify this semantic closeness by leveraging the embeddings of sentences~\cite{zhang2019bertscore,pillutla2021mauve}. Due to the  
% For this purpose, consider \(f: R \rightarrow \mathbb{R}^n\) as a function that maps each response in \(R\) to a continuous vector space, yielding an \(n\)-dimensional embedding.

% Given two responses \(r_1, r_2 \in R\), their respective embeddings can be expressed as \(f(r_1)\) and \(f(r_2)\). The similarity between these responses is then computed as the normalized dot-product of their embeddings:
% \begin{equation}
%     \text{{sim}}(r_1, r_2) = \frac{f(r_1) \cdot f(r_2)}{\|f(r_1)\|_2 \cdot \|f(r_2)\|_2}
% \end{equation}
% where \(\cdot\) denotes the dot-product operation, and \(\|\cdot\|_2\) is the L2 norm of a vector. This equation yields a similarity score in the range [0, 1], where a score of 1 denotes identical semantic content, and a score of 0 implies orthogonal or maximally dissimilar embeddings. 

\begin{figure}[t]
\vspace{-0.2cm}
    \begin{centering}
    \includegraphics[width=1.0\linewidth]{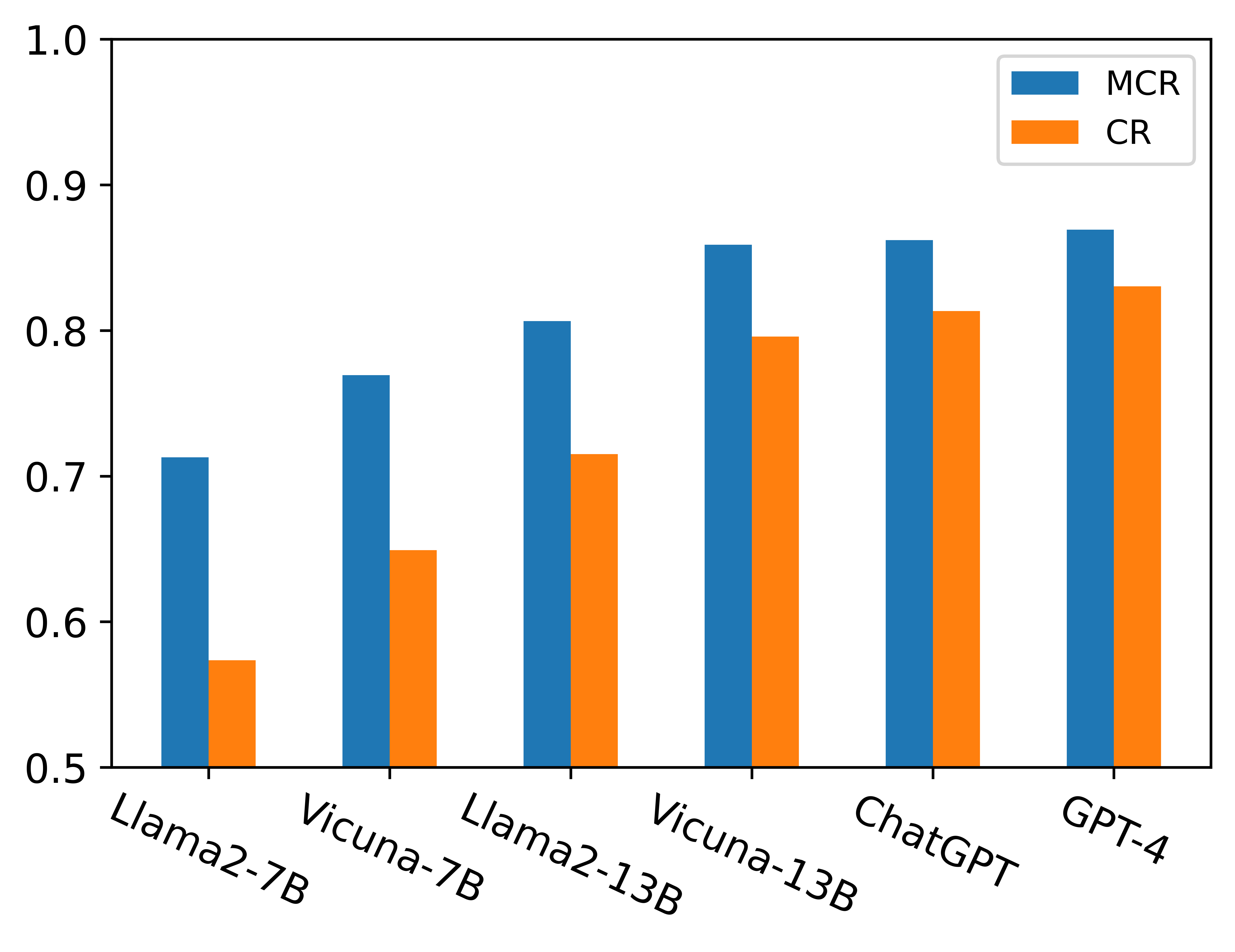}
    \end{centering}
    \vspace{-0.4cm}
    \caption{The consistency metrics of recent LLMs.}
    \label{fig:cr_llms}
    \vspace{-0.4cm}
\end{figure}

\paragraph{Robustness of the current LLMs}

We conduct a preliminary study analyzing the current state of contemporary LLMs quantitatively, namely GPT-3.5, GPT-4, Vicuna and LLaMA-2 in terms of the previously defined metrics.

We crafted a test set randomly sampled 490 questions from Super Natural Instructions~\cite{wang2022super}, each with 10 different linguistic paraphrases, resulting in a total of 4900 unique questions. These questions spanned diverse topics, including science, literature, mathematics, and general knowledge. In our setting, We use GPT-4 to verify the consistency among any two responses. We report the $CR$ and $MCR$ in the Figure~\ref{fig:cr_llms}.
We see GPT-4, ChatGPT and Vicuna-13B emerged as the more robust model in terms of the two consistency metrics. There is still room to improve the robustness especially the smaller ones when contending with diverse linguistic representations. The characteristics of the inconsistency are discussed in the following section. These two metrics are straightforward and can be used to illustrate how far the current LLMs are from the optimal robustness performance.

\begin{figure*}[ht]
\subfigure[Instruction Augmented Supervised Fine-Tuning]{
\includegraphics[width=0.495\linewidth]{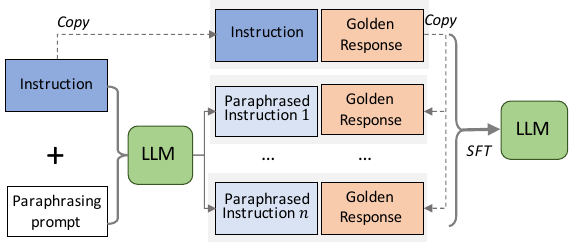}
%\end{minipage}%
%\begin{minipage}[t]{1.0\linewidth}
}
\subfigure[Response Consistency Alignment Training]{
\includegraphics[width=0.495\linewidth]{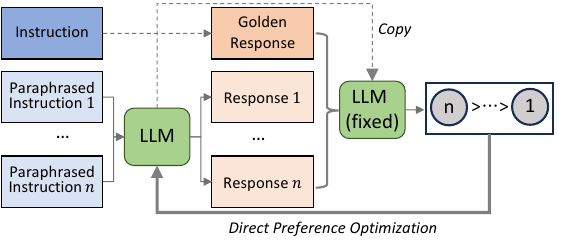}
}
\label{fig:main_framework}
\caption{Our consistency alignment training framework.}
\end{figure*}

% \paragraph{Mutual Information} (alternative)
% Let \(Q\) be a potentially set representing all conceivable linguistic representations with equivalent semantic intent, and let \(O: Q \rightarrow \mathcal{P}(\mathcal{R})\) be a function representing the LLM, mapping each query in \(Q\) to a probability distribution over the possible responses in \(\mathcal{R}\).

% Given this, the robustness \(R\) of an LLM \(O\) with respect to a set of queries \(Q\) is defined as the average mutual information between the distributions of responses to all pairs of queries in \(Q\):

% \begin{equation}
% R(O, Q) = \frac{1}{|Q|^2} \sum_{q_i, q_j \in Q} I(P(O(q_i)); P(O(q_j)))
% \end{equation}

% where \(I(P; Q)\) is the mutual information between two distributions \(P\) and \(Q\), and is defined as:

% \begin{equation}
% I(P; Q) = \sum_{x \in \mathcal{R}} P(x) \log\left( \frac{P(x)}{Q(x)} \right)
% \end{equation}
% Training Framework
\section{Training Large Language Models via Consistency Alignment}
\label{sec:framework}
We aim to improve the robustness of instruction-following for the large language models via consistency alignment. As shown in Figure~\ref{fig:main_framework}, our training framework consists of (1) Supervised fine-tuning with instruction augmentation (SFT (IA)) to improve the model's generalization on following instructions; (2) Consistency alignment training (CAT) with the automatic feedback after the first stage, which helps the model notice diversity and subtle differences between the similar responses rather than simple imitation.

\subsection{Instruction Augmented Supervised Fine-Tuning}
\label{subsec:paraphrase}
Firstly, we augment the task instructions with similar ones to guide the model's instruction tuning.

\paragraph{Instruction Augmentation} Similar instructions are the instructions that convey the same task but are verbalized differently. This aligns with real scenarios where the same task is likely to be induced by different end-users with varying textual expressions. Unfortunately, there is an absence of scaled human-written similar instruction datasets. We prompt the large language models to paraphrase the original instructions into several re-phrasings. The language models are not restricted here where can be Vicuna, ChatGPT or GPT-4\footnote{We have examined the precision of paraphrasing performance for Vicuna-7B, ChatGPT and GPT-4, the precision values are 93
\%, 95\% and 95\% respectively.}. In our paper, there is an original task instruction $a$ along with several input-output instances $M = \{(x_i, y_i)\}$ for the task. We paraphrase the task instruction $a$ and keep the input and output instances unchanged. The prompt we use for the paraphrasing task is shown in Table~\ref{tab:paraphrase_prompt}. %Figure~\ref{fig:paraphrase_prompt} in Appendix~\ref{sec:appendix}. 

\begin{table}[!th]
%\centering
\small
\vspace{-0.1cm}
% \begin{tabular}{l}
\begin{tabular}[c]{p{7.25cm}}
\toprule
\texttt{Paraphrase the input sentences to have different words and expressions but have the same meaning as the original sentences. Output the various paraphrases separated by '<br>'. Please note that you should not answer the question, but rather paraphrase it.} \\
\bottomrule
\end{tabular}
\caption{The instruction for the paraphrasing task.}
\label{tab:paraphrase_prompt}
\vspace{-0.2cm}
\end{table}

After the paraphrasing process, we obtain at most $n$ similar task instructions for each task.

\paragraph{Supervised Fine-Tuning (SFT)} Then we use the paraphrased instructions along with its original instruction to fine-tune our model.
For each task, we randomly sample $m$ instances for the supervised fine-tuning stage. The training set we use is $S= \bigcup_{k} \bigcup_{j}^{n} \bigcup_{i}^{m} \{a_{j}^k, x_i^k, y_i^k\}$, where $a_{j}^k$ is the $j_{th}$ task instruction and $(x_i^k, y_i^k)$ is the $i_{th}$ input-output pair for task $k$. We combine a task instruction $a$ with an input $x_i$ as a question $q_i$ for the model and use $y_i$ as the target output for training.
The training objective in our paper is a standard supervised fine-tuning loss shown below:
\begin{equation}
    L_{sft} = - \sum_t \log P(y_{i, t} | a, x_i, y_{i<t})
\end{equation}

\subsection{Response Consistency Alignment Training}
\label{subsec:cat}
We obtain a trained model after the first stage. We aren't aiming to train a model that merely mimics and lacks diversity even if its instruction-following capabilities are improved.
Therefore, we continue to train the model to learn which responses align with human expectations better, using consistency rewards to differentiate the generated responses. We opt an offline model training method to directly optimize the <good, bad> response pairs for its stability and simplicity like~\cite{rafailov2023direct}.

For each input $x_i$, we utilize the trained model to generate $n$ responses in terms of the $n$ task instructions.
We build training pairs among the $n$ responses where each response is scored individually via self rewards.

\paragraph{Self Rewards} We prompt the trained model to give a reward $r_i$ for each generated response ${y_i}$. 
As we analyze the consistency of the current LLMs in the previous section, we define the inconsistency among the answers in terms of the answer type and the correctness. The answer type indicates whether the model understands the task and is the first step for generation. As shown in Figure~\ref{fig:intro_sample}, the model does not provide a referent but a repeat of the number "two" providing a wrong answer type. The correctness adopts a common definition in language models~\cite{kadavath2022language}.
We ask the model to output whether the generated ${y_i}$ is the expected answer type with reward $r_i^{T} \in \{0, 1\}$, and whether the answer is correct with reward $r_i^{C} \in \{0, 1\}$.
The instructions we used to verify the expected answer type and its correctness are shown in the first and second row in Table~\ref{tab:prompt_reward} respectively. We concat the verification instruction along with the question $q_i$ (i.e., $a$ \text{+} $x_i$), the generated response ${y_i}$ and the golden response as the final prompt to obtain the reward from the LLM itself. The $r_i^{T}$ and $r_i^{C}$ are then inferred from the generated contents by using the keywords "\texttt{Unexpected}", "\texttt{Expected}", "\texttt{Incorrect}" or "\texttt{Correct}". These two tasks are easier for LLMs when used to determine the alignment rewards.
We then combine $r_i^{T}$ and $r_i^{C}$ to a final reward $r_i$ as follows:
\begin{equation}
    r_i = \begin{cases}
0, & r_i^{T}=0 \\
1, & r_i^{T}=1 \wedge r_i^{C}=0 \\
2, & r_i^{T}=1 \wedge r_i^{C}=1 \\
\end{cases}
\end{equation}
% $$ $$

\begin{table*}[htbp]
\small
\begin{tabular}[c]{p{15.5cm}}
\toprule
\texttt{Determine whether the answer "A" is the expected answer type for the question "Q". For tasks with fixed golden answers (like sentiment analysis, entailment inference), you need to check whether the answer is the exact one of the expected enums (like "True", "False", "Positive", "Negative", "A", "B", "Contradiction", "Neutral" or "Entailment".) mentioned in the question "Q". For free-form generation tasks (like title generation, question generation, data-to-text generation), you need to check whether the answer is an expected generated title, question, data-to-text description or summarization, etc., as the question "Q" required. You also need to compare with the "Golden A" to determine whether the answer type aligns with the answer type of the "Golden A". If the answer has the same type as the golden answer, give "Expected answer type", otherwise give "Unexpected answer type". Please note that you only need to determine whether it matches the instructed answer type, and do not need to verify whether the answer is correct.
} \\
\midrule
\texttt{Determine whether the answer "A" is "Correct" or "Incorrect" for question "Q". For tasks with fixed golden answers (the answer is limited to a finite set such as "True", "False", "Positive", "Negative", "A", "B", "Contradiction", "Neutral", "Entailment"), you need to check whether the answer exactly matches (equals) the golden answer "Golden A". For free-form generation tasks (the answer is a free-form generation and not unique such as title, question, data-to-text description generation or summarization), you need to check whether the answer describe the same thing as the golden answer, or the answer is fluent, plausible for the question "Q". If the answer is correct, give "Correct", otherwise give "Incorrect" as the result.
} \\
\bottomrule
\end{tabular}
\caption{The instructions for determining whether a response is the expected answer type and whether it is a correct answer.}
\label{tab:prompt_reward}
\vspace{-0.2cm}
\end{table*}%

% \paragraph{Alignment Training}
% build pair
For each input $x_i$, we obtain response pairs $\bigcup_{j,i} {<y_j, y_i>}$ where $r_j > r_i \wedge r_j = 2$ through the above step. We then incorporate these preferences into the consistency training process, making the model learn to generate more aligned responses with higher scores, and reduce the probability of generating responses that have lower rewards. 
Inspired by~\cite{yuan2023rrhf, rafailov2023direct}, we optimize this objective by the ranking loss:
\begin{equation}
L_{rank} = \sum_{r_i < r_j} \max (0, p_i - p_j)
\end{equation}
where $p_i$ is the conditional log probability of $y_i$ under the model.
We also add the cross-entropy loss $L_{sft}$, where the ground-truth $y_i$ here is replaced with the response that have the highest reward $i^{'} = \mathop{\arg\max}\limits_{i}r_i$. The final loss is the sum of the two losses.
\begin{equation}
    L = L_{rank} + \lambda * L_{sft}
\end{equation}
where $\lambda$ is the weight of training the model for imitation.

% \todo{DPO details}

% training loss
% discussion: 更多接近的样本
\section{Experiments}
\label{sec:experiments}

\subsection{Experimental Setup}
\label{subsec:setup}
\paragraph{Dataset} We use the Super Natural Instructions~\cite{wang2022super} as our experimental dataset, which consists of 1600+ diverse NLP tasks each with at least one expert-written instruction along with 3155 average input-output samples. We adopt the original division for the training and testing sets and randomly split the original training set into training and validation sets. Finally, we obtain 700, 128, and 56 tasks for training, validation, and testing. For each task, we sample at most 100 instances for training or testing after augmenting the paraphrases.

For testing, we build one more test set to report the consitency metrics. We construct the test set \uppercase\expandafter{\romannumeral1} by utilizing GPT-4 to paraphrase the original task instruction obtaining 10 task instructions, and randomly sampling 10 instances for each task. The test set \uppercase\expandafter{\romannumeral2} is the original 56 tasks along with 100 randomly selected samples per task.

\paragraph{Models} We train Vicuna-7B, Vicuna-13B, LLama 2-7B and Llama 2-13B to verify the effectiveness of our proposed training method.

\paragraph{Baselines} We compare our training method with the original LLM, the standard supervised fine-tuning (SFT) method, and the off-the-shelf SOTA LLMs including ChatGPT and GPT-4.
For standard SFT, we randomly sample 100 instances for each task thus obtaining 70000 total samples for training. We set learning rate 2e-5, epochs 3, and other hyperparameters as the FastChat~\footnote{https://github.com/lm-sys/FastChat} suggested.

\paragraph{Evaluation Metrics} We evaluate our method with robustness and accuracy metrics: (1) Consistency Rate $CR$ of any two answers between the same task $Q_k$ under the test set \uppercase\expandafter{\romannumeral1}; (2) Maximum Consistency Rate $MCR$ reports the rate of the maximum consistent answers under the test set \uppercase\expandafter{\romannumeral1};
(3) ROUGE-1, ROUGE-L under the test set \uppercase\expandafter{\romannumeral1} and test set uppercase\expandafter{\romannumeral2} .
Besides, we perform the human evaluation by annotating the quality of the generated responses with scores 0, 1, and 2 ourselves. We report the number of wins, ties, and loses across the responses generated by different models.
% \begin{equation}
%     CR = \frac{1}{|Q|} \sum_{Q_k \in Q} \sum_{y_i\in Q_k} \sum_{y_j\in Q_k, j\neq i} \frac{same(y_i, y_j)}{ \tbinom{|Q_k|}{2}}
% \end{equation}
% where $same(y_i, y_j)$ is inferred from prompting the LLM, and $\tbinom{|Q_k|}{2}$ is the number of 2-combinations for the responses $Q_k$ under the same task and the same input. (2) Maximum Consistency Rate ($MCR$) reports the rate of the maximum consistent answers on the $Q$ Tasks. 
% \begin{equation}
%     MCR = \frac{1}{|Q|} \sum_{Q_k \in Q} \frac{|\Omega_{k}^{max}|}{|Q_k|} 
% \end{equation}
% where $\Omega_{k}^{max} = \arg_{j} \max \Omega_{j}$.
%Besides, we sample 100 different generations between our method and the strongest baseline SFT, and annotate the quality with score 0, 1 and 2 ourselves. Besides, we report the win, tie, loss rate on the diff cases.

\paragraph{Implementation Details} For instruction augmentation, we utilize Vicuna-7B, Vicuna-13B, and ChatGPT to paraphrase the original task instructions obtaining at most 30 instructions. We randomly sample $n=10$ instructions and sample 10 instances for each task for the subsequent training. Thus, the scale of training resources is the same as the baseline. For the SFT stage, we train the models using the FastChat with ZeRO-3 on 4 80G A100 GPUs, setting training epochs 3 and the other hyperparameters as the FastChat suggested. For the CAT stage, we revise minor codes in DPO in LLaMA-Efficient\footnote{https://github.com/hiyouga/LLaMA-Efficient-Tuning} to support our training objective. The rewards are inferred from the LLMs themselves for training, i.e., Vicuna-7B, Vicuna-13B, Llama 2-7B, and Llama 2-13B respectively. We train the models with 3 epochs using LoRA~\cite{hu2021lora} and ZeRO-3 on 4 40G A100 GPUs and set $\lambda=1$, $lora\_r=8, lora\_target=q\_proj,k\_proj,v\_proj,o\_proj$ and $lr=1e-5$. The other hyperparameters are the default setting. To fairly compare the effectiveness of training methods, we do not import any additional data in this stage and re-use the training samples at the SFT stage.

\subsection{Main Results}

\begin{table*}[!t]
\centering\small
%\setlength\tabcolsep{5pt}
% p{3.5em}<{\centering}p{3.5em}<{\centering}
\vspace{-0.3cm}
\begin{tabular}
{ lp{4.5em}<{\centering} p{5.5em}<{\centering} p{5.5em}<{\centering} p{5.5em}<{\centering} p{5.5em}<{\centering} 
}
\toprule
& CR & MCR & ROUGE-1 & ROUGE-L\\
\midrule
GPT-4 & 0.8303 & 0.8693 & 0.3870 & 0.3751 \\
ChatGPT & 0.8134 & 0.8620 & 0.3022 & 0.2744 \\
\midrule
%\multicolumn{4}{@{}l}{\emph{Vicuna-7B performance}}\\
Vicuna-7B & 0.6492 & 0.7694 & 0.1385 & 0.1266 \\
Vicuna-7B + SFT &	0.7092 & 0.8123 & 0.3782 & 0.3672 \\
Vicuna-7B + SFT (IA) & 0.7753 & 0.8504 & 0.3894 & 0.3757 \\
Vicuna-7B + SFT (IA) + CAT & 0.8298 & 0.8743 & 0.4187 & 0.4097 \\
\midrule
%\multicolumn{4}{@{}l}{\emph{Vicuna-7B performance}}\\
Vicuna-13B & 0.7959 & 0.8589 & 0.1724 & 0.1596 \\
Vicuna-13B + SFT &	0.8017 & 0.8490 & 0.4028 & 0.3903 \\
Vicuna-13B + SFT (IA) & 0.8267 & 0.8619 & 0.4131 & 0.4014 \\
Vicuna-13B + SFT (IA) + CAT & 0.8390 & 0.8804 & 0.4276 &  0.4185 \\
\midrule
%\multicolumn{4}{@{}l}{\emph{GPT-3.5 performance}} \\
Llama2-7B & 0.5735 & 0.7129 & 0.0637 & 0.0492  \\
Llama2-7B + SFT & 0.7702 &  0.8308 & 0.2682 &  0.2560 \\
Llama2-7B + SFT (IA) & 0.7921 & 0.8475 & 0.2901 &  0.2733 \\
Llama2-7B + SFT (IA) + CAT & 0.8107 & 0.8521 & 0.3012 & 0.2806 \\
\midrule
%\multicolumn{4}{@{}l}{\emph{GPT-3.5 performance}} \\
Llama2-13B & 0.7151 & 0.8065 & 0.0737 & 0.0627  \\
Llama2-13B + SFT &	0.7505 & 0.8180 & 0.3085 & 0.2975 \\
Llama2-13B + SFT (IA) & 0.7589 &  0.8282 & 0.3379 & 0.3280 \\
Llama2-13B + SFT (IA) + CAT & 0.8100 & 0.8601 & 0.3711 & 0.3502 \\
%\midrule
\bottomrule
\end{tabular}
\caption{The overall performance of compared methods and models on the test set \uppercase\expandafter{\romannumeral1}.}
\label{table:overall_performance}
\vspace{-0.3cm}
\end{table*}

We report the consistency metrics $CR$, $MCR$\footnote{We prompt GPT-4 to judge whether two answers are consistent in the testing stage for consistency metrics, and the prompts are those we discussed in section~\ref{sec:definition}.} as well as ROUGE-1 and ROUGE-L on the compared models and training methods on test set \uppercase\expandafter{\romannumeral1} in Table~\ref{table:overall_performance}. We observe a significant improvement in the consistency scores and ROUGE scores for the Vicuna and Llama 2 models after SFT. SFT with Instruction Augmentation (SFT(IA)) results in higher consistency scores and ROUGE scores compared to standard SFT. The ROUGE scores are improved as well which indicates the instruction augmentation can help to generalize on following instructions. Building upon the SFT (IA) model, consistency alignment training (CAT) brings continual improvements in consistency scores and ROUGE scores for all the compared Vicuna and LLama2 models. Vicuna-13B + SFT (IA) + CAT surpasses the SOTA LLM GPT-4 in our setting. These results demonstrate the effectiveness of our training method (SFT (IA) + CAT). 

Besides, we observe that the performance obtained after SFT (IA) and CAT based on Vicuna is superior to that based on Llama 2, confirming the importance of choosing the base model for further training.
%we observe after SFT(IA) and CAT on Vicuna performs better than Llama 2,  improvements in ROUGE-1 and ROUGE-L for vicuna-13B + SFT, surpassing chatGPT and GPT-4. LLama2-7B + SFT also quickly exceeded its 13B version in ROUGE scores, but their consistency scores remained lower. Subsequent experiments with SFT(IA) and CAT show more visible improvements in consistency scores compared to ROUGE scores, highlighting the importance of observing consistency scores for assessing the model robustness. 
%Note that we utlize the GPt-4 to assess the consistency of generated answers. To ensure fairness, we do not report the consistency scores for chatGPT.

We report the ROUGE-1 and ROUGE-L scores on the standard test set \uppercase\expandafter{\romannumeral2} shown in Table~\ref{table:overall_performance_2}. The improvements and conclusions across the compared methods are consistent as those observed in test set \uppercase\expandafter{\romannumeral1}. Since the Vicuna-13B achieves the best performance, we use it as the backbone for detailed analysis in the next section.

\begin{table}[!t]
\centering\small
\begin{tabular}
{lp{5.0em}<{\centering} p{5.0em}<{\centering}
}
\toprule
& ROUGE-1 & ROUGE-L\\
\midrule
GPT-4 & 0.4506 & 0.4408 \\
ChatGPT & 0.3187 & 0.3051 \\
\midrule
Vicuna-7B & 0.1702 & 0.1570 \\
+SFT & 0.4085 & 0.3929 \\
+SFT (IA) & 0.4122 & 0.3984 \\
+SFT (IA) + CAT & 0.4391 & 0.4285 \\
\midrule
Vicuna-13B & 0.2102 & 0.1972 \\
+SFT & 0.4234 & 0.4071 \\
+SFT (IA) & 0.4477 &  0.4350 \\
+SFT (IA) + CAT & 0.4683 & 0.4417 \\
\midrule
Llama2-7B & 0.0684 & 0.0513  \\
+SFT & 0.2743 & 0.2614 \\
+SFT (IA) & 0.3163 & 0.2903 \\
+SFT (IA) + CAT & 0.3189 & 0.2977 \\
\midrule
Llama2-7B & 0.0745 &  0.0643  \\
+SFT & 0.3351 & 0.3215 \\
+SFT (IA) & 0.3697 & 0.3587 \\
+SFT (IA) + CAT & 0.4289 & 0.4066 \\
%\midrule
\bottomrule
\end{tabular}
\vspace{-0.2cm}
\caption{The ROUGE scores of the compared models and methods on the test set \uppercase\expandafter{\romannumeral2}.}
\label{table:overall_performance_2}
\vspace{-0.6cm}
\end{table}

\subsection{Detailed Analysis}

\paragraph{The Choice of Rewards}
We study whether it is necessary to split the rewards into two steps: one for determining whether it is the expected answer type and the next for correctness. We report the ROUGE values for vicuna-13B on test set \uppercase\expandafter{\romannumeral1} when we construct alignment training pairs using only the correctness reward $r^{C}$ or a combination of correctness and answer type rewards $r^{C} + r^{T}$ in Table~\ref{tab:reward_infulence}. We observe that the ROUGE values have been improved when combining the rewards together, as determining whether it is the expected answer type is an easier task which serves as an auxiliary task to enhance the accuracy of the self rewards. Besides, we study the impact of rewards from different models, i.e., the vanilla Vicuna-13B and the fine-tuned Vicuna-13B with instruction augmentations. We see that using the rewards from a less aligned model would degrade the CAT performance and its ROUGE values are even lower than its SFT version shown in Table~\ref{table:overall_performance_2}.

\begin{table}[t]
\centering
\small
% \vspace{-0.1cm}
\begin{tabular}{@{}lcc@{}}
\toprule
Rewards & ROUGE-1 & ROUGE-L \\
\midrule 
$r^{C}$ from SFT & 0.4123 & 0.4051  \\ 
$r^{C} + r^{T}$ from SFT & 0.4276 & 0.4185 \\ 
$r^{C} + r^{T}$ from Vicuna-13B & 0.3962 & 0.3877 \\ 
 \bottomrule
\end{tabular}
\vspace{-0.2cm}
\caption{The ROUGE scores on test set \uppercase\expandafter{\romannumeral1} when the rewards are expected type or expected type + answer correctness from Supervised fine-tuned Vicuna-13B and vanilla Vicuna-13B.}
\label{tab:reward_infulence}
\vspace{-0.4cm}
\end{table}

\begin{figure}[t]
\vspace{-0.0cm}
    \begin{centering}
    \includegraphics[width=1.0\linewidth]{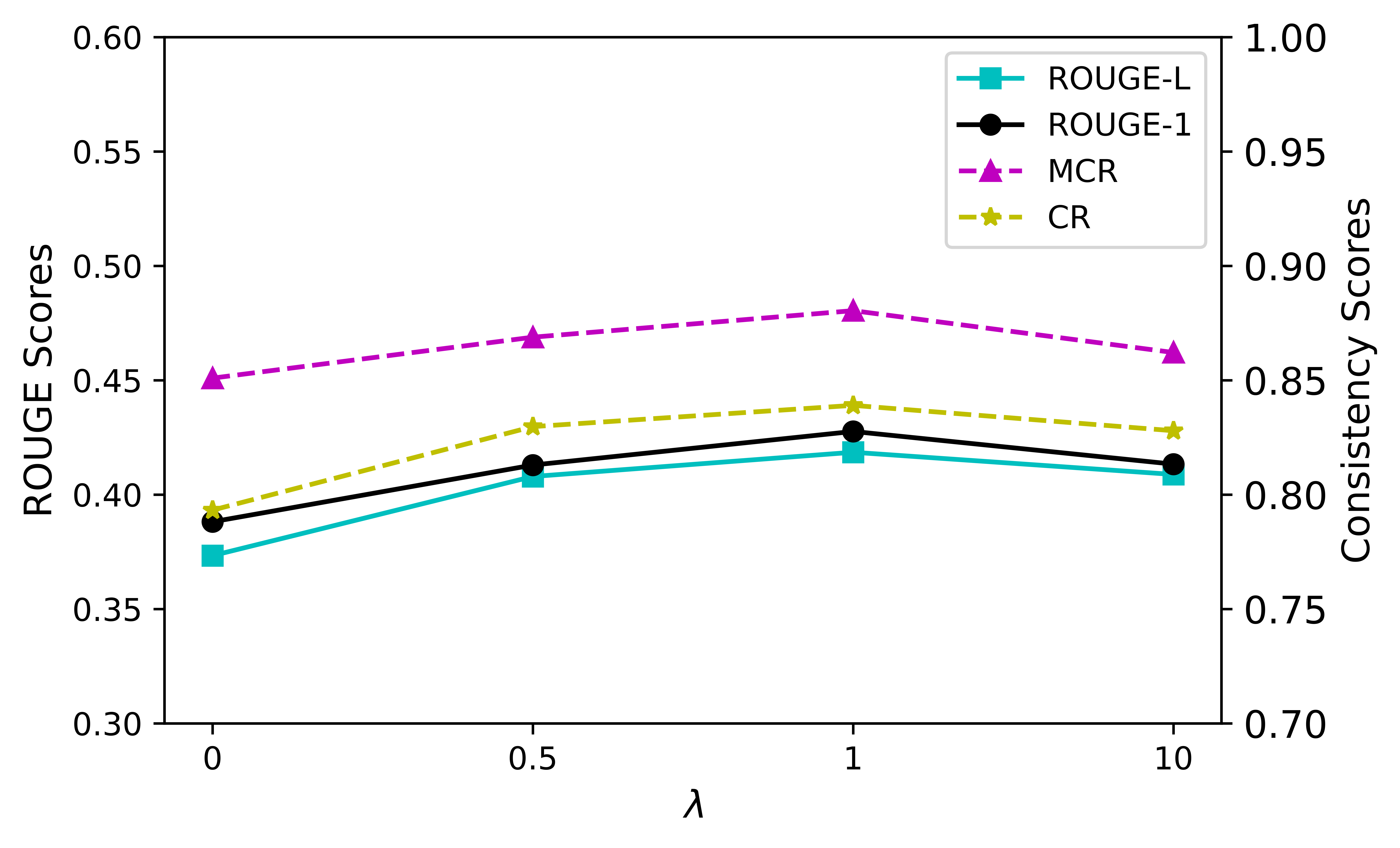}
    \end{centering}
    \vspace{-0.4cm}
    \caption{The performance of different $\lambda$ for our training method on the test set \uppercase\expandafter{\romannumeral1} .}
    \label{fig:diff_lambda}
    \vspace{-0.4cm}
\end{figure}

\paragraph{The Choice of $\lambda$}
We compare the performance of different coefficients of the SFT loss in the final training objective. We report the ROUGE-1 and ROUGE-L scores for Vicuna-13B on test set \uppercase\expandafter{\romannumeral1} when $\lambda$ is set to 0, 0.5, 1, and 10 in Figure~\ref{fig:diff_lambda}. We notice the importance of adding the SFT loss to the CAT objective as the ROUGE scores are higher when we set $\lambda$ to 1 or 0.5 compared with 0.
When we further increase $\lambda$, we make the model learn less from negative generations and reduce the benefits of consistency alignment training. This would decrease the performance in terms of the ROUGE scores.

\begin{figure}[t]
\vspace{-0.2cm}
    \begin{centering}
    \includegraphics[width=1.0\linewidth]{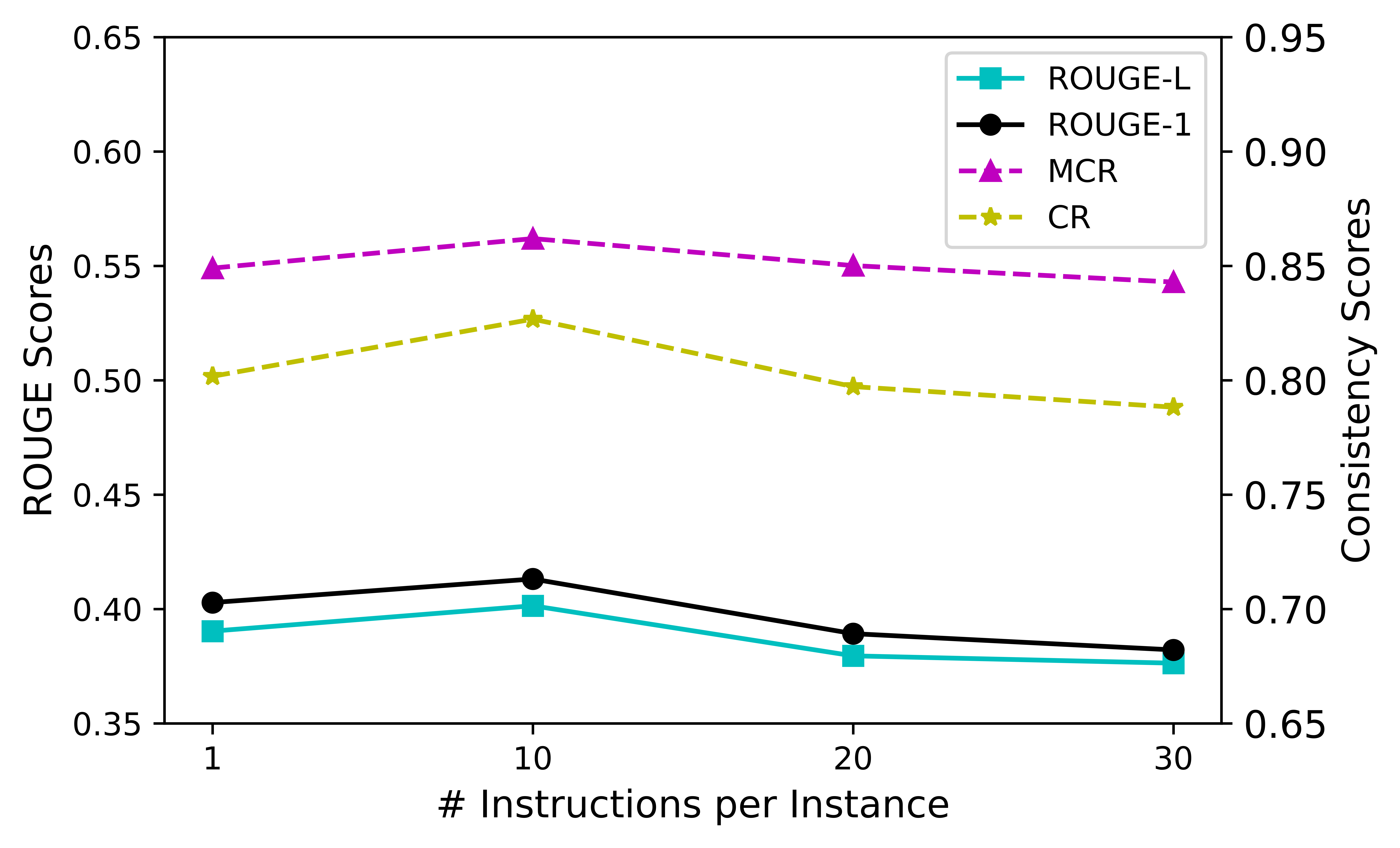}
    \end{centering}
    \vspace{-0.4cm}
    \caption{The performance of SFT (IA) across varying number of instructions for each input on the test set \uppercase\expandafter{\romannumeral1} .}
    \label{fig:diff_aug}
    % \vspace{-0.4cm}
\end{figure}

\paragraph{The Number of Augmented Instructions}
We further study whether the performance would be improved when we increase the number of augmented instructions. For fairness, we keep the size of the training set and other hyperparameters fixed and only increase the number of augmented instructions. We report the metrics of the fine-tuned Vicuna-13B with 1, 10, 20, and 30 instructions per task on test set \uppercase\expandafter{\romannumeral1} in Figure~\ref{fig:diff_aug}. We see that when we increase the number of instructions to 10, both the generated consistency and ROUGE metrics are improved. However, continually increasing the number of instructions would degrade the performance as the number of training instances for each task decreases accordingly.
This encourages that we need to ensure each task is sufficiently trained instead of consistently increasing the number of instructions.

\subsection{Human Evaluation}
We perform human evaluation on the generated responses across different trained models. To compare any two trained models, we sample the different responses (not exactly match, abbreviated as "diff.") generated by the two models in test set \uppercase\expandafter{\romannumeral1}, and we then manually evaluate the pair of these responses\footnote{We evaluate whether a response is an expected answer type and its correctness and score 0, 1, 2 as we discussed in section\ref{subsec:cat}. The number of wins, ties, and loses are inferred by comparing the two scores.}. Based on the statistics in Table~\ref{tab:human_eval}, the performance has been significantly improved through CAT + SFT (IA) compared with its vanilla version, with more wins than loses and a greater diff. ratio. We analyze the human-labeled scores when reporting wins or loses to gain a deeper understanding of how the wins happen.
The ratios of responses scored 1 or 2 and 2 are 86\% and 44\% respectively for CAT + SFT (IA), and the ratios are only 56\% and 14\% for vanilla, verifying the model has been trained to generate more expected answer types and more correct answers.
When comparing CAT + SFT (IA) with SFT, the gap between wins and loses indicates the superior performance of CAT + SFT (IA).
We see that the model trained with consistency alignment after SFT (IA) can obtain consistent improvements compared with SFT (IA) in terms of more wins than loses. 
This indicates that the model can be continually improved to generate more expected answers with additional CAT.

From all the above human evaluated results, we demonstrate that
CAT + SFT(IA) training method helps to generate more aligned responses than the other trained methods.

%87\% of the responses scored 1 or 2 and 45\% scored 2 generated by CAT + SFT (IA) in the diff responses, while the rates for SFT(IA) are 80\% and 33\%. 

\begin{table}[t]
\centering
\small
% \vspace{-0.1cm}
\begin{tabular}{@{}llcccc@{}}
\toprule
Strategy & Baseline & diff. & win & tie & lose \\
\midrule 
CAT+SFT(IA) & Vanilla & 83 & 48 & 44 & 8  \\ 
CAT+SFT(IA) & SFT & 46 & 32 & 55 & 13  \\ 
CAT+SFT(IA) & SFT(IA) & 27 & 31 & 60 & 9  \\ 
 \bottomrule
\end{tabular}
\vspace{-0.1cm}
\caption{Human evaluation on test set \uppercase\expandafter{\romannumeral1}. We report the different ratio, the number of wins, ties and loses among the responses generated by the strategy and baseline models. All the models are trained based on Vicuna-13B and Vanilla denotes the vanilla Vicuna-13B.}
\label{tab:human_eval}
\vspace{-0.1cm}
\end{table}

% \begin{table*}[htbp]
% \toprule
% \texttt{
% Instruction: In this task, you are given triplets. Each triplet is in the form of [subject, predicate, object]. Your task is to generate proper sentence that utilizes these triples. The objective is to construct a sentence that (a) captures the facts specified in the triples and (b) is a well-formed sentence easily understandable by a human. All triple values need not be used directly in the sentence as long as the facts are adequately captured. \\
% Input: [['Northwestern College', 'NICKNAME', 'Red Raiders'], ['Northwestern College', 'LOCATION', 'Orange City, Iowa']] \\
% Response (CAT+SFT): Northwestern College is located in Orange City, Iowa. The college's nickname is Red Raiders. \\
% Response (SFT): Northwestern College is located in Orange City, Iowa.
% }\\
% %\midrule
% \bottomrule
% \caption{The generated response from different trained models.}
% \label{tab:prompt_reward}
% \vspace{-0.2cm}
% \end{table*}%
\section{Conclusion}
\label{sec:conclusion}
In this paper, we investigate the robustness of current large language models in terms of the consistency of the generated responses. We introduce a novel training framework to boost the robustness of LLMs including instruction-augmented supervised fine-tuning (SFT (IA)) and response consistency alignment training (CAT). We conduct extensive experiments on Vicuna and Llama 2 on the instruction-following tasks and validate the effectiveness of our training framework. Additionally, we separately verify the effectiveness of the SFT (IA) and CAT modules.
The method proposed in this paper serves as a plug-in for continuously improving the performance of existing LLMs. Furthermore, it eliminates the need for additional human guidance or external reward models, which use decomposed self-rewards to help the model generate more robust and accurate responses. We believe this approach can contribute to advancing the research on generation robustness to some extent.

% \newpage

\section{limitations}
% 只适用于有一定能力的大模型，更weak的小模型
While our consistency alignment training is effective for improving the robustness and generalization of following instructions, it has several limitations. Firstly, our training approach relies on a model's self-rewards, so if the performance of the model's alignment is poor, the rewards we obtain may not lead to further improvements. In the future, we plan to conduct experiments on smaller and weaker large language models. Besides, the diversity of the verbalized instructions may be limited compared with end-users as we collect the re-phrasings through LLMs. We plan to collect instructions from a wide range of end-users to construct test and training datasets to evaluate and improve the robustness of model generation.

\section{Ethics Statement}
This work does not have explicit ethical considerations as all the models and datasets we used are public. We acknowledge that the LLMs may encode problematic biases. It is unclear how the training process might interact with these problems.

\section{Acknowledgements}
 
This work was supported by 
the National Key R\&D Program of China with grant No.2020YFB1406704, 
the Natural Science Foundation of China (62272274, 61902219, 61972234),
the Natural Science Foundation of Shandong Province (ZR2021QF129). 

%\newpage

% \section{Acknowledgements}

% Place all acknowledgments (including those concerning research grants and funding) in a separate section at the end of the paper.

% \section{Providing References}

% \subsection{Bibliographical References} 

% Bibliographical references should be listed in alphabetical order at the end of the paper. The title of the section, ``Bibliographical References'', should be a Level 1 Heading. The first line of each bibliographical reference should be justified to the left of the column, and the rest of the entry should be indented by 0.35 cm.

% The examples provided in Section~\ref{sec:reference} (some of which are fictitious references) illustrate the basic format required for papers in conference proceedings, books, journal articles, PhD theses, and books chapters.

% \subsection{Language Resource References}

% Language resource references should be listed in alphabetical order at the end of the paper.

\nocite{*}
\section{Bibliographical References}\label{sec:reference}

\bibliographystyle{lrec-coling2024-natbib}
\bibliography{lrec-coling2024-example}

% \section{Language Resource References}
% \label{lr:ref}

\bibliographystylelanguageresource{lrec-coling2024-natbib}
\bibliographylanguageresource{languageresource}

% \newpage
% \appendix
% \section{Appendix}
% \label{sec:appendix}
% \subsection{The prompts in Our Method}

% \begin{figure*}[!ht]
%     \centering
%     \includegraphics[width=1.0\linewidth]{figs/paraphrase_prompt.pdf}
%     \caption{The prompt for paraphrasing similar instructions.}
%     \label{fig:paraphrase_prompt}
%     \vspace{-0.4cm}
% \end{figure*}

\end{document}